\documentclass[nonblindrev]{informs3} 

\usepackage{balance}
\usepackage{amsmath}
\usepackage[tight,footnotesize]{subfigure}
\usepackage{graphicx}
\usepackage{bm}
\usepackage{algorithm}
\usepackage[noend]{algorithmic}
\usepackage{multirow}
\usepackage{slashbox}
\usepackage{caption}
\usepackage{amsfonts}

\OneAndAHalfSpacedXII 

\usepackage{natbib}
 \bibpunct[, ]{(}{)}{,}{a}{}{,}%
 \def\bibfont{\small}%
\TheoremsNumberedThrough     

\EquationsNumberedThrough    

\MANUSCRIPTNO{2015} 

\begin{document}


\RUNAUTHOR{Tang et al.}

\RUNTITLE{Optimal Sampling Gaps for Adaptive Submodular Maximization}

\TITLE{Optimal Sampling Gaps for Adaptive Submodular Maximization}

\ARTICLEAUTHORS{%
\AUTHOR{Shaojie Tang}
\AFF{Naveen Jindal School of Management, The University of Texas at Dallas}
\AUTHOR{Jing Yuan}
\AFF{Department of Computer Science, The University of Texas at Dallas}
} 

\ABSTRACT{Running machine learning algorithms on large and rapidly growing volumes of data is often computationally expensive, one common trick to reduce the size of a data set, and thus reduce the computational cost of machine learning algorithms, is \emph{probability sampling}. It creates a sampled data set by including each data point from the original data set with a known probability. Although the benefit of running machine learning algorithms on the reduced data set is obvious, one major concern is that the performance of the solution obtained from samples might be much worse than that of the optimal solution when using the full data set.  In this paper, we examine the performance loss caused by  probability sampling in the context of adaptive submodular maximization. We consider a simple probability sampling method which selects each data point with probability at least $r\in[0,1]$.  If we set $r=1$, our problem reduces to finding a solution based on the original full data set. We define sampling gap as the largest ratio between the optimal solution obtained from the full data set and the optimal solution obtained from the samples,  over independence systems. 
Our main contribution is to show that if the sampling probability of each data point is at least $r$ and the utility function is policywise submodular, then the sampling gap is both upper bounded and lower bounded by $1/r$. 
We show that the property of policywise submodular can be found in a wide range of real-world applications, including pool-based active learning and adaptive viral marketing.}


\maketitle

\section{Introduction}

Many machine learning methods are highly benefitted when they are fed with  the right volume of data. One common approach to reduce the volume of a large data set is probability sampling, which generates a sampled data set by including each data point with a known probability.  However, one major concern of running an algorithm on a sampled data set is that the performance of the sampling-based solution might be much worse than that of the optimal solution when using the full data set.  In this paper, we examine the performance loss caused by  probability sampling in the context of adaptive submodular maximization over independence systems.

Due to the wide applicability of submodular functions, submodular maximization, whose objective is to select a group of items to maximize a submodular function on various types of independence systems, including matroid \citep{nemhauser1978analysis,calinescu2007maximizing} and knapsack \citep{sviridenko2004note}, has been extensively studied in the literature.  Most of existing studies focus on the non-adaptive setting,  where each item has a deterministic state and all items must be selected at once. However, the classic notation of submodularity can not capture the interactive nature of many applications, including active learning and experimental design, where one must adaptively select a group of items based on the stochastic observations collected from past selections. Recently, \citep{golovin2011adaptive}  introduce the notation of adaptive submodularity, which extends the classic notation of submodularity from sets to policies. In the problem of adaptive submodular maximization, each item has a particular state which is unknown initially. One must pick an item before observing its realized state. An adaptive policy can be represented using a decision tree which specifies which item to pick next based on the realizations observed so far. 

\cite{golovin2011adaptive} develop a simple adaptive greedy algorithm that achieves a $1-1/e$ approximation for the problem of cardinality constrained adaptive submodular maximization. Their algorithm starts with an empty set, and in each iteration, it selects an item with the largest marginal utility on top of the current observation. This algorithm requires $n\times k$ value oracle queries, where $n$ is the size of the ground set and $k$ is the cardinality constraint. However, evaluating the marginal utility of an item is expensive in many data intensive applications, making the standard greedy algorithm infeasible in practise. One natural idea to reduce the computational cost of any machine learning algorithms is to run them on a reduced ground set that is sampled from the full set. One major concern is that the output restricted to the sampled data set might be much worse than that of the optimal solution when using the full data set. This raises the \emph{sampling gap} question:

\emph{ What is the maximum ratio between the expected utility of the optimal solution (over independence systems) when using the full data set and that when using the sampled data set?}

If this sampling gap is small then we can focus on finding a good solution based on the sampled data set while enjoying its benefits of reduced computational cost and small performance loss.  In this work, we consider a simple sampling method that selects each item from the full set with probability at least $r\in[0,1]$.  Our objective is to examine the performance loss due to the probability sampling in the context of adaptive submodular maximization.   

\textbf{Overview of Results.} We first introduce a class of stochastic functions, called \emph{policywise submodular function}. Policywise submodularity refers to the property of diminishing returns over optimal policies, and we show that this property can be found in a wide range of real-world applications, including pool-based active learning and adaptive viral marketing. Our main contribution is to show that if the sampling probability of each item is at least $r$ and the utility function is policywise submodular, then the sampling gap, i.e., the maximum ratio between the optimal solution  based on the full data set and the optimal solution  based on samples, over independence systems, is both upper bounded and lower bounded by $1/r$. One major implication of our result is that if we can find an $\alpha$-approximation solution based on a sampled data set such that each item is being sampled with probability at least $r$, then this solution achieves an $\alpha r$ approximation ratio against the optimal solution when using the full data set.   

\section{Related Works}
The adaptive variant of submodular maximization  has been extensively studied in the literature \citep{chen2013near,tang2020influence,tang2020price,yuan2017adaptive,fujii2019beyond,gabillon2013adaptive,golovin2010near,alaei2021maximizing}. For the case of maximizing an adaptive monotone and adaptive submodular function subject to a cardinality constraint, \cite{golovin2011adaptive} develops a simple adaptive greedy policy that achieves a tight $1-1/e$ approximation ratio. For maximizing a nonmonotone adaptive submodular function, \cite{tang2021beyond,tang2021mco-beyond} develop the first constant approximation algorithms.  
Given the rapid growth of data volume, much recent research in submodular maximization has explored the possibility of developing fast and practical algorithms \citep{leskovec2007cost,badanidiyuru2014fast,mirzasoleiman2016fast,ene2018towards,mirzasoleiman2015lazier,tang2021beyond,feldman2017greed}, and many of them have adopted the technique of random sampling to develop better approximation algorithms or reduce the computational cost for maximizing (non-adaptive) submodular functions subject to various constraints. The other line of workstudies the problem of maximizing  (non-adaptive) monotone submodular functions from samples \citep{balkanski2016power}. Unlike our setting where the utility function (of the sampled data set) is known, they assume that the function optimized is not known a priori, it is learned from data.
Summarizing, all aforementioned results focus on the non-adaptive setting and their goal is to investigate the impact of random sampling on maximizing submodular functions subject a particular algorithm or a particular constraint. 
Our study complements the existing studies by establishing a general framework for measuring the performance loss of the optimal solution caused by the probability sampling in the context of adaptive submodular maximization. Our results are not restricted to any particular algorithms or constraints, and the utility function is not necessarily monotone and it might take on negative values.

\section{Preliminaries}
We start by introducing some important notations. In the rest of this paper, 
we use $|X|$ to denote the cardinality of a set $X$.

\subsection{Independence System}

An \emph{independence system} $\mathcal{I}$ on the set $V$ is a collection of subsets of $V$ such that:
\begin{enumerate}
\item $\emptyset \in \mathcal{I}$;
\item $\mathcal{I}$, which is called the \emph{independent sets}, is downward-closed, that is, $A \in \mathcal{I}$ and $B \subseteq A$ implies that $B \in \mathcal{I}$. 
\end{enumerate}

Examples of independence systems include matroid, knapsack, matching, and independent set. The \emph{upper rank} $\textrm{rank}( \mathcal{I})$ of an independence system $\mathcal{I}$ on $V$ is defined as the size of the largest subset from  $\mathcal{I}$, i.e., $\textrm{rank}( \mathcal{I})=\max_{I\in \mathcal{I}} |I|$. For any two sets $S\in \mathcal{I}$ and $R\subseteq V$ such that $S\cap R = \emptyset$,  let $\mathcal{I}^{R}_{S} = \{A| A\cup S \in \mathcal{I}, A\subseteq R\}$.

We next present three useful properties of any independence system $(V, \mathcal{I})$. These properties will be used later to derive the main results of this paper. All missing proofs are deferred to the full version \citep{tang2021optimal}.
\begin{lemma}
\label{lem:6}
Consider any two sets $S\in \mathcal{I}$ and $R\subseteq V$ such that $S\cap R = \emptyset$, we have $(V, \mathcal{I}^{R}_S)$ is an independence system and $\mathcal{I}^{R}_S \subseteq \mathcal{I}$.
\end{lemma}

\begin{lemma}
\label{lem:4}
Assume  $\textrm{rank}( \mathcal{I})>0$. For any item $e\in V$ such that  $\{e\} \in \mathcal{I}$, we have $(V, \mathcal{I}^{V\setminus\{e\}}_{\{e\}})$ is an independence system and $\textrm{rank}(\mathcal{I}^{V\setminus\{e\}}_{\{e\}}) < \textrm{rank}(\mathcal{I})$.
\end{lemma}

\begin{lemma}
\label{lem:3}
Consider any three sets $S\in \mathcal{I}$, $S'\in \mathcal{I}$, and $R\subseteq V$ such that $S'\subseteq  S$ and $S\cap R = \emptyset$, we have $\mathcal{I}^{R}_S \subseteq \mathcal{I}^{R}_{S'}$.
\end{lemma}

\subsection{Items and  States} We consider a set  $V$ of $n$ items, where each item is in a particular state, which is unknown initially, from  $O$.  Each item $e\in V$ has  a random state $\Phi(e) \in O$. Let $\phi(e) \in O$ denote a realization of $\Phi(e)$. Thus, a \emph{realization} $\phi$ is a mapping function that maps items to states: $\phi: V \rightarrow O$. In the example of experimental design, the item $e$ may represent a test, such as the heart rate, and
$\Phi(e)$ is the outcome of the test, such as, $70$ per minute.   
There is a known prior probability distribution $p(\phi)=\{\Pr[\Phi=\phi]: \phi\in O^V\}$ over realizations. Note that when realizations are independent, the distribution $p$ completely factorizes. However, this independent assumption may not hold in many real-world applications such as experimental design and active learning. We assume that one must select an item  $e\in V$ before observing  the value of $\Phi(e)$. For any subset of items $S\subseteq V$, we use $\psi: S\rightarrow O$ to denote a \emph{partial realization} and $\mathrm{dom}(\psi)=S$ is called the \emph{domain} of $\psi$. A partial realization $\psi$ is said to be consistent with a realization $\phi$, denoted $\phi \sim \psi$, if they are equal everywhere in $\mathrm{dom}(\psi)$. A partial realization  $\psi$  is said to be a \emph{subrealization} of another partial realization $\psi'$, denoted  $\psi \subseteq \psi'$, if $\mathrm{dom}(\psi) \subseteq \mathrm{dom}(\psi')$ and they are equal everywhere in the domain $\mathrm{dom}(\psi)$ of $\psi$. Given a partial realization $\psi$, denote by $p(\phi\mid \psi)$ the conditional distribution over realizations conditioned on  $\psi$: $p(\phi\mid \psi) =\Pr[\Phi=\phi\mid \Phi\sim \psi ]$.

\subsection{Policies} Any adaptive policy can be represented as a function $\pi$ that maps a set of observations  to a distribution $\mathcal{P}(V)$ of $V$: $\pi: 2^{V}\times O^V  \rightarrow \mathcal{P}(V)$. It specifies  which item to select next based on  the past outcomes.
%
%
%
There is a utility function $f: 2^{V}\times O^V\rightarrow \mathbb{R}$ from a subset of items and their states to a real number.

\begin{definition}[Conditional Marginal Utility of an Item]
\label{def:1kkk}
Given a utility function $f: 2^{V}\times O^V\rightarrow \mathbb{R}$, the conditional expected marginal utility of any item $e\in V$ on top of any partial realization $\psi$ is defined as
$f_{avg}(e \mid \psi)=\mathbb{E}_{\Phi}[f(\mathrm{dom}(\psi)\cup \{e\}, \Phi)-f(\mathrm{dom}(\psi), \Phi)\mid \Phi \sim \psi]$,
where the expectation is taken over $\Phi$ with respect to $p(\phi\mid \psi)$.
\end{definition}

Let the random variable $V(\pi, \phi)$ denote the subset of items selected by a policy $\pi$ under realization $\phi$.  The expected  utility $f_{avg}(\pi)$ of a policy $\pi$  can be written as
\begin{eqnarray*}
f_{avg}(\pi)=\mathbb{E}_{\Phi\sim p(\phi), \Pi}f(V(\pi, \Phi), \Phi)
\end{eqnarray*}
The expectation is taken over the realization and the internal randomness of the policy.

\begin{definition}[Conditional Marginal Utility of a Policy]
\label{def:1}
Given a utility function $f: 2^{V}\times O^V\rightarrow \mathbb{R}$, the conditional expected marginal utility $f_{avg}(\pi \mid \psi)$ of a policy $\pi$ on top of a partial realization $\psi$ is
$f_{avg}(\pi\mid \psi)=\mathbb{E}_{\Phi, \Pi}[f(\mathrm{dom}(\psi) \cup V(\pi, \Phi), \Phi)-f(\mathrm{dom}(\psi), \Phi)\mid \Phi\sim \psi]$, where the expectation is taken over realizations $\Phi$ with respect to $p(\phi\mid \psi)$ and the internal randomness of the policy.

\end{definition}

\begin{definition}[$\mathcal{I}^{R}_{S}$-restricted Policy]
\label{def:1}
Consider two sets $S\in \mathcal{I}$ and $R\subseteq V$ such that $S\cap R = \emptyset$. A policy $\pi$ is $\mathcal{I}^{R}_{S}$-restricted if for all $\phi$ such that $p(\phi)>0$, we have $V(\pi, \phi) \in \mathcal{I}^{R}_{S}$. Let $\Omega(\mathcal{I}^{R}_{S})$ denote the set of all $\mathcal{I}^{R}_{S}$-restricted polices, i.e., $\Omega(\mathcal{I}^{R}_{S})=\{\pi\mid \mbox{ for all $\phi$ such that $p(\phi)>0$}: V(\pi, \phi) \in \mathcal{I}^{R}_{S}\}$.
\end{definition}

\begin{definition}[Optimal $\mathcal{I}^{R}_{S}$-restricted Policy on top of $\psi$]
\label{def:1}
Consider two sets $S \in \mathcal{I}$ and $R\subseteq V$ such that $S\cap R = \emptyset$, and a partial realization $\psi$. Define $\pi^*(\mathcal{I}^{R}_{S}, \psi)$ as the optimal $\mathcal{I}^{R}_{S}$-restricted policy on top of $\psi$, i.e.,
\begin{eqnarray*}
\pi^*(\mathcal{I}^{R}_{S}, \psi) \in \arg\max_{\pi\in \Omega(\mathcal{I}^{R}_{S})} f_{avg}(\pi| \psi)
\end{eqnarray*}
\end{definition}

With the above notation, $\pi^*(\mathcal{I}^V_\emptyset, \emptyset)$ represents the optimal policy over an independence system $(V, \mathcal{I})$, i.e., $\pi^*(\mathcal{I}^V_\emptyset, \emptyset) \in \arg\max_{\pi\in \Omega(\mathcal{I}^V_\emptyset)} f_{avg}(\pi|\emptyset)$ or equivalently, $\pi^*(\mathcal{I}^V_\emptyset, \emptyset) \in \arg\max_{\pi\in \Omega(\mathcal{I}^V_\emptyset)} f_{avg}(\pi)$.

\subsection{Policywise Submodularity and Sampling Gap}
In this paper, we propose a new class of stochastic utility functions, \emph{policywise submodular functions}.  We define policywise
submodularity as the diminishing return property about the expected marginal gain of the optimal policy over independence systems.
\begin{definition}[Policywise Submodularity]
A function  $f: 2^{V}\times O^V\rightarrow \mathbb{R}$ is policywise submodular with respect to a prior $p(\phi)$ and an independence system  $(V, \mathcal{I})$ if for any two partial realizations $\psi^a$ and $\psi^b$ such that $\psi^a\subseteq \psi^b$ and $\mathrm{dom}(\psi^b)\in \mathcal{I}$, and any $R\subseteq V$ such that $R\cap \mathrm{dom}(\psi^b) =\emptyset$, we have
\begin{eqnarray*}
f_{avg}(\pi^*(\mathcal{I}^{R}_{\mathrm{dom}(\psi^b)}, \psi^a)| \psi^a) \geq f_{avg}(\pi^*(\mathcal{I}^{R}_{\mathrm{dom}(\psi^b)}, \psi^b)| \psi^b)
\end{eqnarray*}
\end{definition}

Later we show that this property can be found in many real-world applications. For example, a variety of objective functions, including generalized binary search \citep{golovin2011adaptive}, EC$^2$ \citep{golovin2010near}, ALuMA \citep{gonen2013efficient}, and the maximum Gibbs error criterion \citep{cuong2013active}, used in active learning are policywise submodular. This property can also be found in other applications wherever the states of the items are independent \citep{asadpour2016maximizing}. Moreover, we prove that the utility function of the adaptive viral marketing \citep{golovin2011adaptive} is also policywise submodular.

 Next we introduce the concept of \emph{sampling gap}, which is defined as the ratio of the optimal solution obtained from the full data set and the optimal solution obtained from the sampled data set, over independence systems. The sampling gap measures the performance loss of the optimal solution due to the probability sampling. 
 In the rest of this paper, for any $R\subseteq V$, let $\pi^*_R$ denote $\pi^*(\mathcal{I}^R_\emptyset, \emptyset)$ for short, i.e., $\pi^*_R$ represents the optimal policy that selects items only from $R$. Hence, $\pi^*_V$ represents the optimal policy using the full data set.
\begin{definition}[Sampling Gap]
Let $T$ be a random subset of $V$ drawn from a distribution $\mathcal{D}$ such that each item is included in $T$ with probability at least $r \in [0, 1]$ (not necessarily independently). 
Define the sampling gap at minimum rate $r$ as the largest (worst-case instance of $(f, p(\phi), V, \mathcal{I}, \mathcal{D})$) ratio of the optimal  policy when using the full ground set $V$ and optimal  policies when using the sampled set $T$, i.e.,
\begin{eqnarray*}
\verb"sampling gap" = \max_{(f, p(\phi), V, \mathcal{I}, \mathcal{D})}\frac{ f_{avg}(\pi^*_V)}{\mathbb{E}_T[f_{avg}(\pi^*_T)]}
\end{eqnarray*}
\end{definition}

In this paper, we restrict our attention to the case of maximizing a policywise submodular function over independence systems. I.e., our goal is to provide an answer to the following question:

\emph{What is the value of $\max_{(f, p(\phi), V, \mathcal{I}, \mathcal{D})}\frac{ f_{avg}(\pi^*_V)}{\mathbb{E}_T[f_{avg}(\pi^*_T)]}$ given that $f: 2^{V}\times O^V\rightarrow \mathbb{R}$ is policywise submodular with respect to $p(\phi)$ and  $(V, \mathcal{I})$?}

\section{Sampling Gaps for a Policywise Submodular Function}
In this section we provide our main result, the optimal sampling gap for policywise submodular functions over independence systems. We prove the upper bound  and lower bound of Theorem \ref{thm:main} in the following two subsections respectively.

\begin{theorem}
\label{thm:main} The sampling gap at minimum rate $r$ for maximizing a policywise submodular function over independence systems is exactly $1/r$.
\end{theorem}
\subsection{Upper Bound of $1/r$}
\label{sec:1}
We first present an upper bound of the sampling gap over independence systems. Note that for an
arbitrary partial realization $\psi$ such that $\mathrm{dom}(\psi)\in \mathcal{I}$, Lemma \ref{lem:6} implies that $(V, \mathcal{I}^{V\setminus\mathrm{dom}(\psi)}_{\mathrm{dom}(\psi)})$ is an independence system. Before presenting the main theorem, we first prove a useful technical lemma. We introduce a new function $f( \cdot,\cdot | \psi): 2^{V}\times O^V\rightarrow \mathbb{R}$ for any partial realization $\psi$ such that $f(S, \phi | \psi) = f(S \cup \mathrm{dom}(\psi), \phi) - f(\mathrm{dom}(\psi),\phi)$ for any $S\subseteq V$ and realization $\phi$ such that $\phi \sim \psi$.
\begin{lemma}
\label{lem:5} Suppose $f: 2^{V}\times O^V\rightarrow \mathbb{R}$ is policywise submodular with respect to a prior $p(\phi)$ and an independence system  $(V, \mathcal{I})$. For an
arbitrary partial realization $\psi$ such that $\mathrm{dom}(\psi)\in \mathcal{I}$,  $f( \cdot | \psi): 2^{V}\times O^V\rightarrow \mathbb{R}$ is also policywise submodular with respect to a prior $p(\phi|\psi)$ and an independence  system  $(V, \mathcal{I}^{V\setminus\mathrm{dom}(\psi)}_{\mathrm{dom}(\psi)})$.\end{lemma}

Now we are ready to present the main theorem of this paper. For convenience, let $f(\emptyset)$ denote the expected utility of an empty set for short, i.e., $f(\emptyset) = \mathbb{E}_{\Phi\sim p(\phi)}[f(\emptyset, \Phi)]$.
\begin{theorem}
\label{thm:1}Let $T$ be a random subset of $V$ where each item is included in $T$ with probability $r \in [0, 1]$ (not necessarily independently).  Suppose $f: 2^{V}\times O^V\rightarrow \mathbb{R}$ is policywise submodular with respect to a prior $p(\phi)$ and an independence system  $(V, \mathcal{I})$,
\[\mathbb{E}_T[f_{avg}(\pi^*_T)] \geq (1-r)f(\emptyset) + rf_{avg}(\pi^*_V)\]
\end{theorem}
\emph{Proof:} We prove this lemma through the induction on the upper rank $\textrm{rank}(\mathcal{I})$ of the independence system  $(V, \mathcal{I})$.

For the base case when $\textrm{rank}(\mathcal{I}) =0$, we have $f_{avg}(\pi^*_T)=f(\emptyset)$ for any $T$ and $f_{avg}(\pi^*_V)=f(\emptyset)$. Hence, $\mathbb{E}_T[f_{avg}(\pi^*_T)] =f(\emptyset)\geq (1-r)f(\emptyset) + rf(\emptyset)= (1-r)f(\emptyset) + rf_{avg}(\pi^*_V)$.

Assume the statement holds for all independence systems $(V, \mathcal{I})$ such that $\textrm{rank}(\mathcal{I}) \leq l-1$, we next prove that it holds for all independence systems $(V, \mathcal{I})$ when $\textrm{rank}(\mathcal{I}) =l$.  To avoid the trivial case we assume that $\pi^*_V$ selects at least one item. Assume $s\in V$ is the root of the decision tree of $\pi^*_V$, i.e., $s$ is the first item selected by $\pi^*_V$, we next construct a policy $\pi_T$ such that
\begin{itemize}
\item If $s\in T$, $\pi_T$ first selects $s$, then adopts  $\pi^*(\mathcal{I}^{T\setminus\{s\}}_{\{s\}},\Phi(s))$ which is the optimal $\mathcal{I}^{T\setminus\{s\}}_{\{s\}}$-restricted policy on top of $\Phi(s)$.
\item If $s\notin T$, $\pi_T$ adopts $\pi^*(\mathcal{I}^{T}_{\{s\}},\emptyset)$ which is the optimal $\mathcal{I}^{T}_{\{s\}}$-restricted policy   on top of $\emptyset$.
\end{itemize}

We first show that $\pi_T$ is a feasible $\mathcal{I}^{T}_\emptyset$-restricted policy.
\begin{lemma}
\label{lem:1}
$\pi_T$ is a feasible $\mathcal{I}^{T}_\emptyset$-restricted policy.
\end{lemma}

Because $\pi^*_T$ is an optimal  $\mathcal{I}^{T}_\emptyset$-restricted policy and $\pi_T$ is a feasible $\mathcal{I}^{T}_\emptyset$-restricted policy (Lemma \ref{lem:1}), we have $f_{avg}(\pi^*_T)\geq f_{avg}(\pi_T)$ for any $T$. Hence, $\mathbb{E}_{T}[f_{avg}(\pi_T)]\leq \mathbb{E}_T[f_{avg}(\pi^*_T)]$. To prove this theorem, it suffices to show that
\begin{eqnarray}
\label{eq:12}
\mathbb{E}_{T}[f_{avg}(\pi_T)]\geq (1-r)f(\emptyset) + r f_{avg}(\pi^*_V)
 \end{eqnarray}

Hence, we next focus on proving (\ref{eq:12}).
%
We first compute the expected utility of $\pi_T$ for a given $T$. In the case of $s\in T$, we have

\begin{eqnarray}
&&f_{avg}(\pi_T) = f(\emptyset) + f_{avg}(s|\emptyset) + \mathbb{E}_{\Phi(s)}[f_{avg}(\pi^*(\mathcal{I}^{T\setminus\{s\}}_{\{s\}},\Phi(s))| \Phi(s) )] \label{eq:3}
\end{eqnarray}

In the case of $s\notin T$, we have

\begin{eqnarray}
f_{avg}(\pi_T) &=& f(\emptyset) +f_{avg}(\pi^*(\mathcal{I}^{T}_{\{s\}},\emptyset)|\emptyset)\geq f(\emptyset) +\mathbb{E}_{\Phi(s)}[f_{avg}(\pi^*(\mathcal{I}^{T}_{\{s\}},\Phi(s))|\Phi(s))] \label{eq:2}
\end{eqnarray}
The inequality is due to $f: 2^{V}\times O^V\rightarrow \mathbb{R}$ is policywise submodular with respect to  $p(\phi)$ and  $(V, \mathcal{I})$. Taking the expectation over $T$, we next bound the expected utility of $\pi_T$.

\begin{eqnarray}
&&\mathbb{E}_{T}[f_{avg}(\pi_T)] = \nonumber\\
&&r\Big( f(\emptyset) + f_{avg}(s|\emptyset) \nonumber\\
&&+ \mathbb{E}_{\Phi(s), T}[f_{avg}(\pi^*(\mathcal{I}^{T\setminus\{s\}}_{\{s\}},\Phi(s))| \Phi(s))| s\in T]\Big)+(1-r)\left( f(\emptyset)  +\mathbb{E}_{T}[f_{avg}(\pi^*(\mathcal{I}^{T}_{\{s\}},\emptyset)|\emptyset)|s\notin T]\right)\nonumber\\
&\geq& r\Big( f(\emptyset) + f_{avg}(s|\emptyset) + \mathbb{E}_{\Phi(s), T}[f_{avg}(\pi^*(\mathcal{I}^{T\setminus\{s\}}_{\{s\}},\Phi(s))| \Phi(s))| s\in T ]\Big)\nonumber\\
&& + (1-r) \Big(f(\emptyset) +\mathbb{E}_{\Phi(s), T}[f_{avg}(\pi^*(\mathcal{I}^{T}_{\{s\}},\Phi(s))|\Phi(s))| s\notin T]\Big)\nonumber\\
&=& r\Big( f(\emptyset) + f_{avg}(s|\emptyset) + \mathbb{E}_{\Phi(s), T}[f_{avg}(\pi^*(\mathcal{I}^{T\setminus\{s\}}_{\{s\}},\Phi(s))| \Phi(s))]\Big)\nonumber\\
&& + (1-r) \Big(f(\emptyset) +\mathbb{E}_{\Phi(s), T}[f_{avg}(\pi^*(\mathcal{I}^{T\setminus\{s\}}_{\{s\}},\Phi(s))|\Phi(s))]\Big)\nonumber\\
&=& (1-r)f(\emptyset)+ r\left(f(\emptyset) + f_{avg}(s|\emptyset)\right)+ \mathbb{E}_{\Phi(s), T}[f_{avg}(\pi^*(\mathcal{I}^{T\setminus\{s\}}_{\{s\}},\Phi(s))| \Phi(s))]\nonumber\\
&\geq& (1-r)f(\emptyset) + r\Big(f(\emptyset) + f_{avg}(s|\emptyset)+ \mathbb{E}_{\Phi(s)}[f_{avg}(\pi^*(\mathcal{I}^{V\setminus\{s\}}_{\{s\}},\Phi(s))|\Phi(s))]\Big)\nonumber\\
&=& (1-r)f(\emptyset) + r f_{avg}(\pi^*_V) \label{eq:4}
\end{eqnarray}
The first inequality is due to (\ref{eq:3}) and (\ref{eq:2}). The second inequality is due to the observation that for any $\Phi(s)$,

\begin{eqnarray}
&&\mathbb{E}_{T}[f_{avg}(\pi^*(\mathcal{I}^{T\setminus\{s\}}_{\{s\}},\Phi(s))| \Phi(s))]\geq  rf_{avg}(\pi^*(\mathcal{I}^{V\setminus\{s\}}_{\{s\}},\Phi(s))|\Phi(s))\label{eq:13}
\end{eqnarray}
(\ref{eq:13}) follows from the inductive assumption based on the following three facts: (1) $\textrm{rank}(\mathcal{I}^{V\setminus\{s\}}_{\{s\}}) < \textrm{rank}(\mathcal{I})=l$ (Lemma \ref{lem:4}), (2) $f( \cdot \mid \Phi(s))$ is policywise submodular with respect to a prior $p(\phi|\Phi(s))$ and an independence system  $(V, \mathcal{I}^{V\setminus\{s\}}_{\{s\}})$ for any $\Phi(s)$ (Lemma \ref{lem:5}), and (3) $f_{avg}( \emptyset \mid \Phi(s)) = 0$  for any $\Phi(s)$. $\Box$

It is easy to verify that increasing the sampling probability of any items does not decrease the expected utility of an optimal policy restricted to a sampled data set. Hence, Theorem \ref{thm:1} implies the following corollary:
\begin{corollary}\label{cor:1}
Given that $f: 2^{V}\times O^V\rightarrow \mathbb{R}$ is policywise submodular with respect to $p(\phi)$ and  $(V, \mathcal{I})$, and $f(\emptyset)\geq0$, the sampling gap at minimum rate $r$ is upper bounded by $1/r$.
\end{corollary}

\emph{Remark:} A similar result was provided in \citep{fujii2016budgeted} for the case when the utility function is policy-adaptive submodular (Definition \ref{def:12}). Our result is more general than theirs because policy-adaptive submodularity is a strictly stronger condition than policywise submodularity (Lemma \ref{lem:7}).

The following corollary shows that if we can find an approximation  solution over a sampled data set, then this solution achieves a bounded approximation ratio for the original problem when using the full data set.
\begin{corollary}
\label{cor:2}
Let $T$ be a random subset of $V$ where each item is included in $T$ with probability at least $r \in [0, 1]$ (not necessarily independently).  If $f: 2^{V}\times O^V\rightarrow \mathbb{R}$ is policywise submodular with respect to a prior $p(\phi)$ and an independence system  $(V, \mathcal{I})$, and there exists an $\alpha$-approximation $\mathcal{I}^{T}_\emptyset$-restricted policy $\pi^\alpha_T$ for every $T$, i.e., $\forall T, \exists \pi^\alpha_T\in\Omega(\mathcal{I}^{T}_\emptyset), f_{avg}(\pi^\alpha_T) \geq \alpha f_{avg}(\pi^*_T)$, then $\frac{ f_{avg}(\pi^*_V)}{\mathbb{E}_T[f_{avg}(\pi^\alpha_T)]} \leq \frac{1}{\alpha r}$.
\end{corollary}

\subsection{Lower Bound of $1/r$}
\label{sec:2}
In this section we construct a policywise submodular function and a simple cardinality constraint to show that the sampling gap at minimum rate $r$ is lower bounded by $1/r$.  Assume $V=\{e\}$, i.e., the ground set contains only one item, and $O=\{o\}$, i.e., there is only one state. Define $f(\emptyset, (e, o))=0$ and $f(\{e\},  (e, o))=1$. Moreover, $\mathcal{I}=\{\{e\}, \emptyset\}$. We first show that $f: 2^{V}\times O^V\rightarrow \mathbb{R}$  is policywise submodular with respect to $p(\phi)$ and $\mathcal{I}$. Because $\mathcal{I}^{\emptyset}_{\{e\}}=\{\emptyset\}$, we have $f_{avg}(\pi^*(\mathcal{I}^{\emptyset}_{\{e\}}, \emptyset)|\emptyset)=f_{avg}(\pi^*(\mathcal{I}^{\emptyset}_{\{e\}}, \{o\})|(e, o))=0$, which implies that $f: 2^{V}\times O^V\rightarrow \mathbb{R}$  is policywise submodular with respect to $p(\phi)$ and $\mathcal{I}$.  Let $T$ be a random set where $e$ appears with probability $r$. In the case of $T=\{e\}$,  $\pi^*_T$ selects $e$, thus, $\mathbb{E}_T[f_{avg}(\pi^*_T)|e\in T]=1$. In the case of $T=\{\emptyset\}$, $\pi^*_T$ selects $\emptyset$, thus, $\mathbb{E}_T[f_{avg}(\pi^*_T)|e\notin T]=0$. Thus, $\mathbb{E}_T[f_{avg}(\pi^*_T)]=r\mathbb{E}_T[f_{avg}(\pi^*_T)|e\in T]+(1-r)\mathbb{E}_T[f_{avg}(\pi^*_T)|e\notin T]=r$. Moreover, because $\pi^*_V$ always selects $e$, $f_{avg}(\pi^*_V)=1$. Hence, the sampling gap at minimum rate $r$ is lower bounded by $1/r$.
\section{Applications}
\label{sec:app}
We next show that the property of policywise submodularity can be found in a wide range of real-world applications. We start with introducing two well-studied classes of stochastic functions. Then we build a relation between these two classes of functions and our notation of policywise submodularity.
\begin{definition}
\label{def:12}\citep{fujii2016budgeted}[Policy-adaptive Submodularity]
A function $f: 2^{V}\times O^V\rightarrow \mathbb{R}$ is policy-adaptive submodular with respect to $p(\phi)$, if for any two partial realizations $\psi^a$ and $\psi^b$ such that $\psi^a\subseteq \psi^b$, and any policy $\pi$ such that  $V(\pi,\phi) \subseteq V\setminus \mathrm{dom}(\psi^b)$ for all $\phi$ such that $p(\phi)>0$, we have $f_{avg}(\pi | \psi^a) \geq f_{avg}(\pi | \psi^b)$.
\end{definition}

\begin{definition}\citep{golovin2011adaptive}[Adaptive Submodularity]
A function $f: 2^{V}\times O^V\rightarrow \mathbb{R}$ is adaptive submodular with respect to $p(\phi)$, if for any two partial realizations $\psi^a$ and $\psi^b$ such that $\psi^a\subseteq \psi^b$, and any item  $e\notin \mathrm{dom}(\psi^b)$, we have $f_{avg}(e | \psi^a) \geq f_{avg}(e | \psi^b)$.
\end{definition}

Our next lemma shows that policy-adaptive submodularity implies both adaptive submodularity and policywise submodularity. Later, we show that policy-adaptive submodularity is a strictly stronger condition than policywise submodularity.
\begin{lemma}
\label{lem:2} If $f: 2^{V}\times O^V\rightarrow \mathbb{R}$ is policy-adaptive submodular with respect to $p(\phi)$, then $f: 2^{V}\times O^V\rightarrow \mathbb{R}$ is both adaptive submodular with respect to $p(\phi)$  and policywise submodular with respect to $p(\phi)$ and any independence system $(V, \mathcal{I})$.
\end{lemma}

Thanks to the recent progress in adaptive submodular maximization \citep{golovin2011adaptive,tang2021beyond}, there exists efficient solutions for maximizing an adaptive submodular function subject to many practical constraints, including matroid and knapsack constraints. Hence, Lemma \ref{lem:2}, together with Corollary \ref{cor:2}, implies that if a function $f: 2^{V}\times O^V\rightarrow \mathbb{R}$  is policy-adaptive submodular with respect to $p(\phi)$, then running existing algorithms on the sampled ground set has comparable performance to running them on the full set.

We next discuss three representative applications whose objective function satisfies the policywise
submodularity. The objective functions in the first two applications are policy-adaptive submodular, which implies both adaptive submodularity and policwise submodularity. Although the objective function in the third application does not satisfy the policy-adaptive submodularity, it is still adaptive submodular and policwise submodular.  This implies that policy-adaptive submodularity is a strictly stronger condition than policywise submodularity.

\paragraph{Application 1: Pool-based Active Learning \citep{golovin2011adaptive}.} We use $\mathcal{H}$ to denote the set of candidate hypothesis. Each
hypothesis $h \in \mathcal{H}$ represents some realization, i.e., $h : V \rightarrow \Phi$, where each item $e\in V$ can be viewed as a data point and the state $\Phi(e)$ of a data point $e$ can be viewed as the label of $e$. Let $p_H$ be a prior distribution over
hypotheses. Define $p_H(\mathcal{H}') = \sum_{h\in \mathcal{H}'} p_H(h)$
for any $\mathcal{H}' \subseteq \mathcal{H}$. Then the prior distribution over realizations can be represented as $p(\phi) = p_H(h| \phi\sim h)$. The version space under observations $\psi$  is defined to be $\mathcal{H}(\psi) = \{h\in \mathcal{H}| h\sim \psi\}$, i.e., $\mathcal{H}(\psi)$ contains all hypothesis whose labels are consistent with $\psi$ in the domain of $\psi$. Given a realization $\phi$ and a group of labeled data points $S$, the utility function of generalized binary search under the Bayesian setting is
  \begin{eqnarray}
f(S, \phi) =  1- p_H(\mathcal{H}(\phi(S))) \label{eq:10}
\end{eqnarray}
where $\phi(S)=\{(e, \phi(e))\mid e\in S\}$.

(\ref{eq:10}) measures the reduction in version space mass after obtaining the states (a.k.a. labels) about $S$. Intuitively, our objective is to select (a.k.a. query the labels of) a group of data points to maximize the expected reduction in version space mass. It has been shown that the above utility function is both adaptive submodular  with respect to $p(\phi)$ (Section 9 in \citep{golovin2011adaptive}) and policy-adaptive submodular with respect to $p(\phi)$ (Proposition A.1 in \citep{fujii2016budgeted}). This, together with Lemma \ref{lem:2}, implies the following proposition.
\begin{proposition}
The utility function $f: 2^{V}\times O^V\rightarrow \mathbb{R}$ of pool-based active learning  is policywise submodular with respect to $p(\phi)$ and any independence system $(V, \mathcal{I})$.
\end{proposition}

Moreover, many other types of objective functions of active learning, including EC$^2$ \citep{golovin2010near}, ALuMA \citep{gonen2013efficient}, and the maximum Gibbs error criterion \citep{cuong2013active}, are both adaptive submodular  and policywise submodular.

\paragraph{Application 2: The Case of Independent Items \citep{asadpour2016maximizing}.} The property of policywise submodularity can also be found in any applications in which the states of items are independent of each other. One such application is sensor selection \citep{golovin2011adaptive}. The following proposition follows from Lemma \ref{lem:2} and the fact that if $\Phi(1), \Phi(2), \dots, \Phi(n)$ are independent and $f$ is adaptive submodular with respect to $p(\phi)$, then $f$ is policy-adaptive submodular with respect to $p(\phi)$ (Proposition A.5 in \citep{fujii2016budgeted}).

\begin{proposition}
If $\Phi(1), \Phi(2), \dots, \Phi(n)$ are independent and $f: 2^{V}\times O^V\rightarrow \mathbb{R}$ is adaptive submodular with respect to $p(\phi)$, then $f: 2^{V}\times O^V\rightarrow \mathbb{R}$ is
policywise submodular  with respect to $p(\phi)$ and any independence system $(V, \mathcal{I})$.
\end{proposition}

\paragraph{Application 3: Adaptive Viral Marketing \citep{golovin2011adaptive}.} The third application is the adaptive variant of viral marketing \citep{golovin2011adaptive}. We use a directed graph $G=(V, E)$ to represent a social network, where $V$  represents a set of individuals and $E$ represents a set of edges. Under the Independent Cascade Model \citep{kempe2008cascade}, each edge $(u,v)\in E$ is associated with a propagation probability $p_{uv}\in[0,1]$. In step $0$, we activate a set of \emph{seeds}. Then, in each subsequence step $t$, each individual $u$, that is newly activated, has a single chance to activate each of its neighbors $v$; it succeeds with a probability $p_{uv}$. If $u$ succeeds, then $v$ becomes activated in step $t+1$. This process is iterated till no more individuals are newly activated. Under the adaptive setting \citep{golovin2011adaptive}, we model the state $\phi(u)$ of $u$ as a function $\phi(u): E\rightarrow \{0, 1, ?\}$, where $\phi(u)((u,v))=0$ means that selecting $u$ reveals that $(u,v)$ is blocked (i.e., $u$ fails to activate $v$), $\phi(u)((u,v))=1$ means that  selecting $u$ reveals that $(u,v)$ is live (i.e., $u$ succeeds in activating $v$), and $\phi(u)((u,v))=?$ means that  selecting $u$ can not reveal the status of $(u,v)$. We assume that selecting a seed $u$ can reveal the status (live or block) of every out-going edge of every individual that can be reached by $u$ though a path composed of live edges. For a given set of seeds $S$ and a realization $\phi$, we define the utility $f(S, \phi)$  as the number of individuals that can be reached by at least one seed from $S$ through live edges (including $S$), i.e.,

{\small\begin{eqnarray}
\label{eq:11}
f(S, \phi) =|\{v| \exists u\in S, w\in V, \phi(u)((w, v))=1 \}|+|S|
\end{eqnarray}}
It has been shown that the above utility function is adaptive submodular (Section 8 in \citep{golovin2011adaptive}). We next show that this function is also policywise submodular.

\begin{proposition}
\label{pro:3}
The utility function $f: 2^{V}\times O^V\rightarrow \mathbb{R}$ of adaptive viral marketing is policywise submodular with respect to $p(\phi)$ and any independence system $(V, \mathcal{I})$.
\end{proposition}

In the proof of the next lemma, we show that (\ref{eq:11}) is not policy-adaptive submodular.
\begin{lemma}
\label{lem:7}
Policy-adaptive submodularity is a strictly stronger condition than policywise submodularity.
\end{lemma}
\begin{figure*}[hptb]
\begin{center}
\includegraphics[scale=0.13]{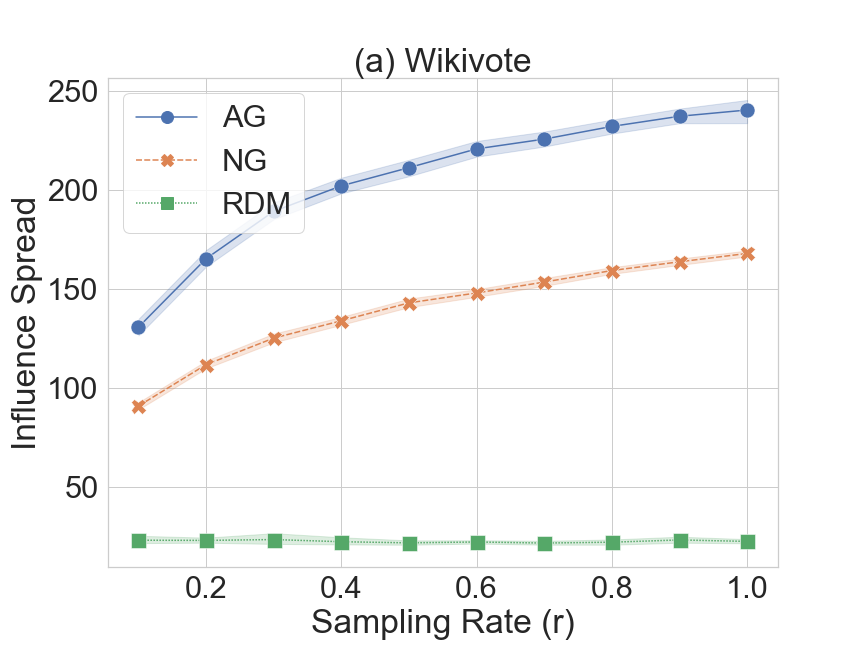}
\includegraphics[scale=0.13]{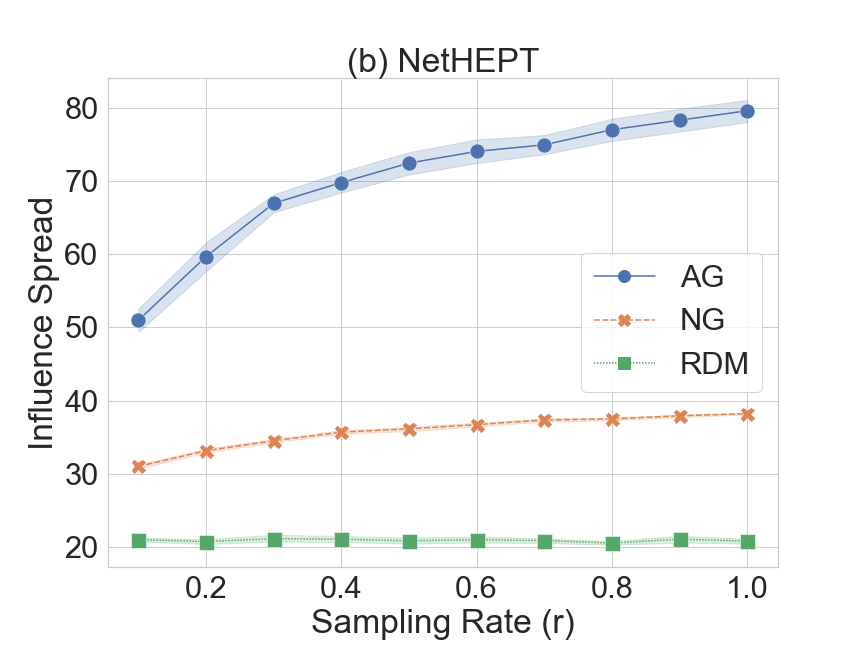}
\includegraphics[scale=0.13]{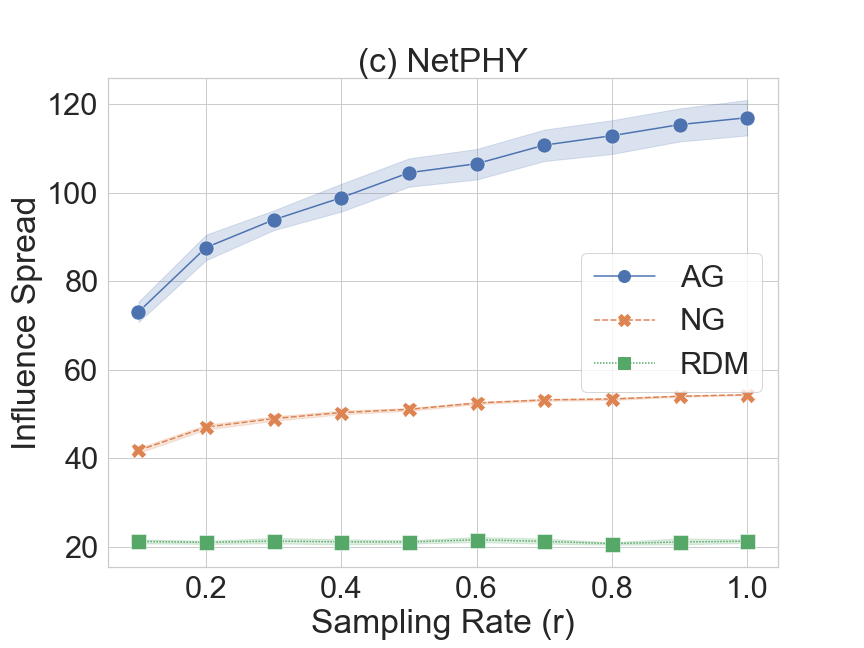}
\includegraphics[scale=0.13]{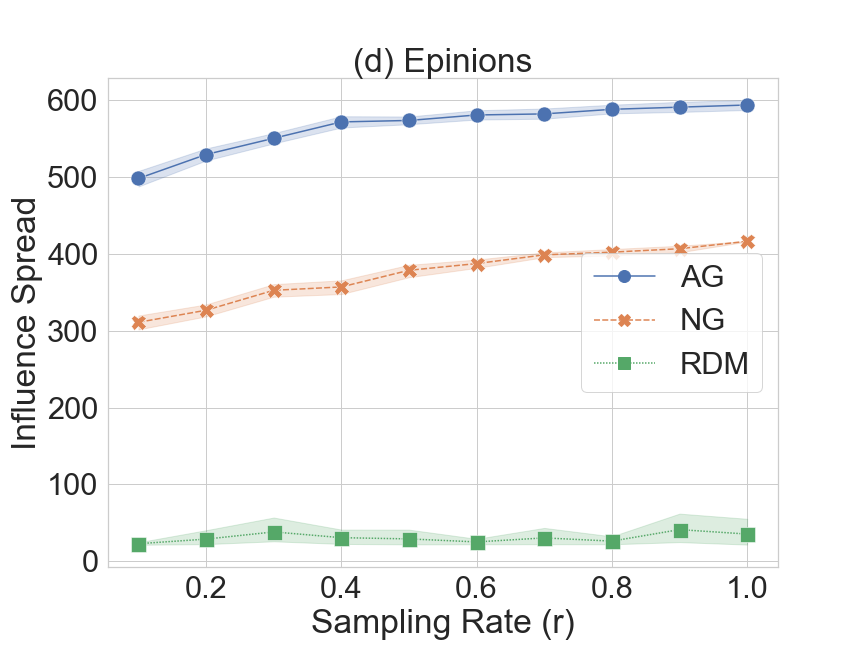}
\caption{Adaptive Viral Marketing: Influence Spread vs. Sampling Rate.}
\label{fig:infmax}
\end{center}
\end{figure*}

\begin{figure*}[hptb]
\begin{center}
\includegraphics[scale=0.13]{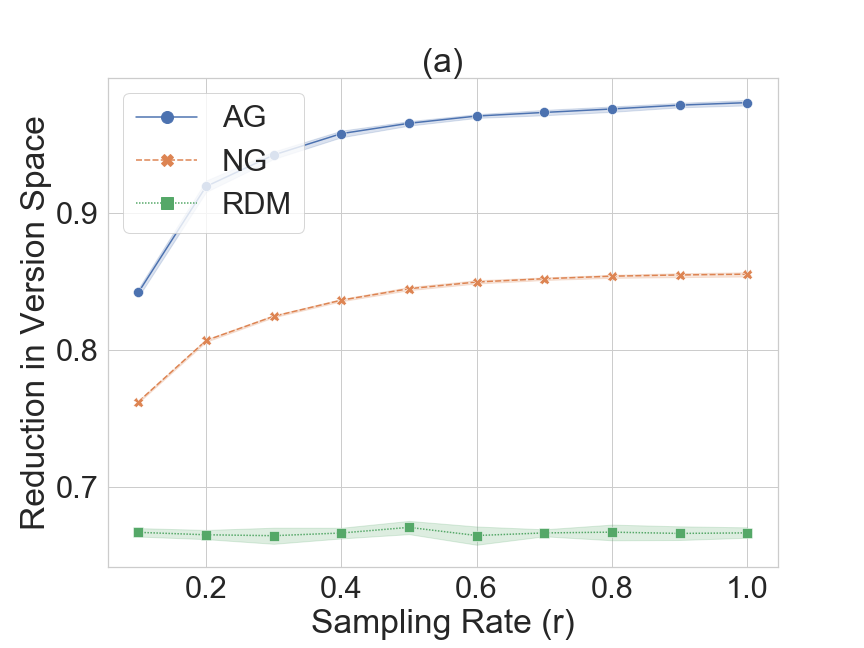}
\includegraphics[scale=0.13]{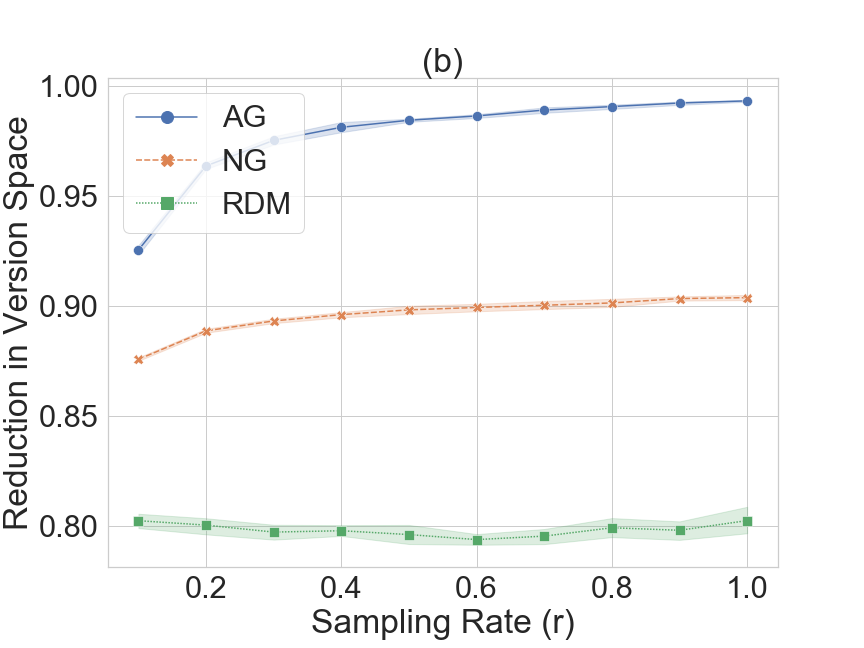}
\includegraphics[scale=0.13]{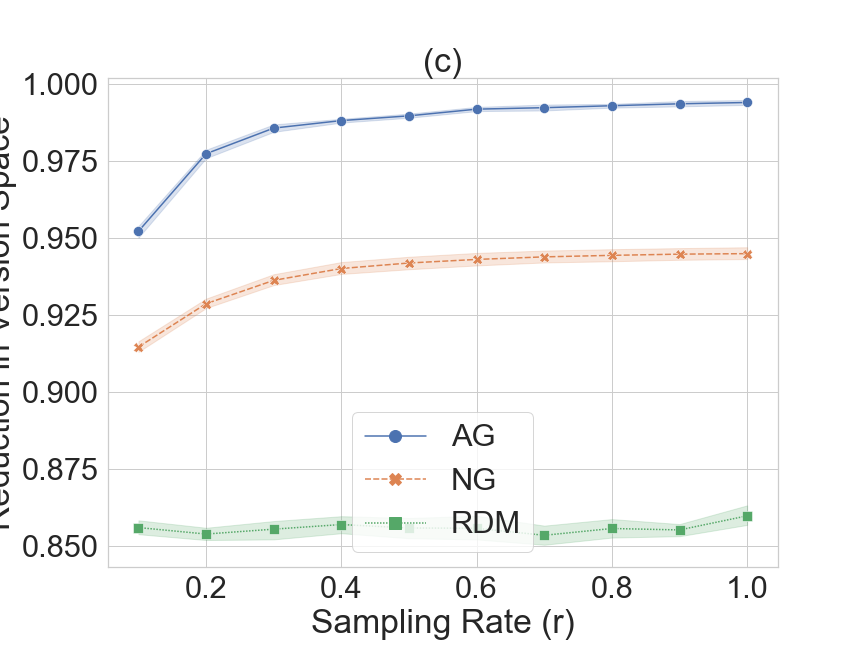}
\includegraphics[scale=0.13]{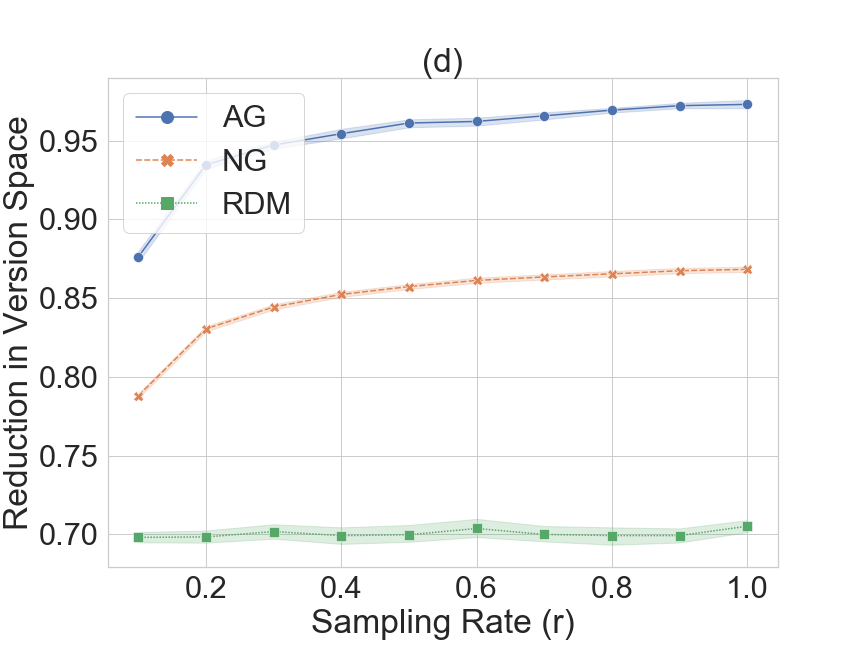}
\caption{Pool-based Active Learning: Reduction in Version Space vs. Sampling Rate}
\label{fig:pbal}
\end{center}
\end{figure*}
\section{Performance Evaluation}
We conduct experiments to evaluate the impact of probability sampling based on two popular machine learning applications: adaptive viral marketing and pool-based active learning \citep{golovin2011adaptive}.


{\bf \emph{Adaptive Viral Marketing.}} A detailed description of this application can be found in Application 3 from the previous section.  We capture the social network by a directed weighted graph and run experiments on four large-scale benchmark social networks: \emph{Wikivote}, \emph{NetHEPT}, \emph{NetPHY} and \emph{Epinions} (http://snap.stanford.edu/data/).  \emph{Wikivote} records $103,663$ votes from $7,066$ users participating in the elections from the Wikipedia community.  \emph{NetHEPT} is a large academic collaboration network extracted from the High Energy Physics Theory section of arXiv, including $15,233$ nodes each representing an author and $62,774$ edges each representing one paper co-authored by two nodes.  \emph{NetPHY} is another academic collaboration network extracted from the Physics section of arXiv, which contains $37,154$ nodes and $231,584$ edges.  \emph{Epinion} is a Who-trust-whom network of Epinions.com, containing $75,879$ nodes and $508,837$ edges. Each node represents a user and each edge represents a trust relationship.  we adopt the Independent Cascade model as diffusion model and assign a uniform probability of $0.01$ to each edge as discussed in \citep{kempe2003maximizing}.

{\bf \emph{Pool-based Active Learning.}}  A detailed description of this application can be found in Application 1 from the previous section. 
We consider $1,000$ hypotheses, $80$ unlabeled data points and $50$ queries each covers one or two unlabeled data points.  The probability of each hypothesis is drawn from $(0,1)$ uniformly at random with normalization; each data point is assigned a value randomly selected from its possible set of labels.  

\hspace*{-0.14in}{\bf Algorithms and Parameters.} For the adaptive viral marketing application, we evaluate the performance of the following three algorithms under various settings: adaptive greedy algorithm (AG), non-adaptive greedy algorithm (NG) and random algorithm (RDM).  AG adaptively selects a node that introduces the largest expected marginal influence spread in each round based on prior observations until $k$ nodes are selected.  Here prior observations refer to the realized influence propagation in the network triggered by the previously selected seeds. Although AG is not necessarily optimal, it can approximate the optimal policy within a factor of $1-1/e$ \citep{golovin2011adaptive}, i.e., the expected size of the influence generated by AG is at least $1-1/e$ times the optimal solution. Therefore, through examining the performance of AG under varies sampling rates, it can help us understand the impact of probability sampling on the optimal solution. The second algorithm NG is a non-adaptive version of AG. It adds the node that maximally increases the objective value before any observations take place.  We run simulations $10,000$ times and took the average in order to obtain reasonable estimates of the influence spread.  For RDM, we randomly select a set of seed nodes of size $k$ as an output.

We sample each node independently at a sampling rate $r$, that is, each node is being sampled independently with probability equal to $r$. We vary $r$ from $0.1$ to $1$. When $r=1$, algorithms run on a full data set that includes all nodes in the network.  A lower $r$ indicates a smaller (in expectation) candidate set from which we select our seed nodes.  
For each sampling rate, we obtain $30$ samples, and report the average performance of each algorithm over these samples, denoted as the point on the line, as well as the confidence interval of the result, denoted as the bar around the line.

For the pool-based active learning application, we evaluate the performance of AG, NG and RDM under various sampling rates.  AG iteratively identifies a query with the highest marginal version space reduction based on prior observations until $5$ queries have been selected.  Here prior observations refer to the user's answer to the query.  NG is a non-adaptive version of AG. RDM randomly selects a set of $5$ queries as an output.  We use a similar sampling setting as above in our experiments.

\hspace*{-0.14in}{\bf Experimental results.}  We present the results in Figure \ref{fig:infmax} and Figure \ref{fig:pbal}. We observe that the performance trends of the algorithms are overall consistent between all datasets and application domains.  For the adaptive viral marketing, we evaluate the performance of algorithms as measured by the influence spread with respect to the changes in the sampling rate on four datasets.   We set $k=10$ for Epinions, and $k=20$ for other three datasets. We observe that among three algorithms examined, AG is the best in all settings.  Surprisingly, on three larger datasets \emph{NetHEPT}, \emph{NetPHY} and \emph{Epinions}, AG with $r=0.1$ outperforms NG with $r=1$, which demonstrates the power of adaptivity.  As expected, RDM has the poorest performance across all settings.  We also notice that when the sampling rate is larger than $0.6$, the performance loss (as compared with when $r=1$) of AG is within $10\%$.  Therefore, we can choose $r=0.6$ to reduce the computational cost significantly without much sacrifice in performance.

For the active learning application, we evaluate the performance of algorithms as measured by the yielded reduction in version space with respect to the changes in the sampling rate. The results are plotted in Figure \ref{fig:pbal}.  We first consider the scenario where each data point has the same number of possible labels. We set the number of labels to $2$, $3$, $4$ and the results are shown in Figure \ref{fig:pbal}(a), (b) and (c) respectively.  To diversity the data points in our data set, we then randomly divide our $80$ unlabeled data points into three groups. The first group contains $64$ data points with binary labels. The second and third group each contains $8$ data points with three and four possible labels respectively. The results are shown in Figure \ref{fig:pbal}(d).  AG is the best under all settings, and its performance tends to stabilize after $r$ is greater than $0.6$.  In particular, the performance loss caused by probability sampling is within $10\%$ for $r\geq 0.3$ and is within $5\%$ for $r\geq 0.5$.  This verifies the value of probability sampling in the context of pool-based active learning.

%
%

\bibliographystyle{ijocv081}
\bibliography{reference}

\newpage
\section*{Appendix}
\subsection{Proof of Lemma \ref{lem:6}}
\emph{Proof:} We first prove that $(V, \mathcal{I}^{R}_S)$ is an independence system.  Consider any two sets $A$ and $B$ such that $A \in \mathcal{I}^{R}_S$ and $B \subseteq A$. Because  $A \in \mathcal{I}^{R}_S$, we have $A\cup S \in \mathcal{I}$ and  $A\subseteq R$. Since  $B \subseteq A$ and $\mathcal{I}$ is downward-closed, we have $B\cup S \in \mathcal{I}$. Moreover, $B \subseteq A$ also implies that $B\subseteq R$. Hence, $B \in \mathcal{I}^{R}_S$, which implies that $\mathcal{I}^{R}_S$ is downward-closed. To show that $\emptyset\in \mathcal{I}^{R}_S$, it suffices to prove that $\emptyset\subseteq  R$ and $\emptyset \cup S \in \mathcal{I}$. The first condition clearly holds and the second condition  is the assumption made in the lemma. We next prove that  $\mathcal{I}^{R}_S \subseteq \mathcal{I}$. For any set $A\in \mathcal{I}^{R}_S$, we have $A\cup S \in \mathcal{I}$, thus, $A \in \mathcal{I}$ due to $\mathcal{I}$ is downward-closed. Hence,  $\mathcal{I}^{R}_S \subseteq \mathcal{I}$. $\Box$

\subsection{Proof of Lemma \ref{lem:4}}
\emph{Proof:} The first part of this lemma, i.e., $(V, \mathcal{I}^{V\setminus\{e\}}_{\{e\}})$ is an independence system, follows from Lemma \ref{lem:6}. We next prove that  $ \textrm{rank}(\mathcal{I})\geq |A\cup\{e\}|> |A|=\textrm{rank}(\mathcal{I}^{V\setminus\{e\}}_{\{e\}})$. Let $A\in \arg\max_{I\in \mathcal{I}^{V\setminus\{e\}}_{\{e\}}} |I|$, i.e., $\textrm{rank}(\mathcal{I}^{V\setminus\{e\}}_{\{e\}})=|A|$.  We have $A\cup\{e\}\in \mathcal{I}$ and $A\cup\{e\}\subseteq V$, thus,  $ \textrm{rank}(\mathcal{I})\geq |A\cup\{e\}|> |A|=\textrm{rank}(\mathcal{I}^{V\setminus\{e\}}_{\{e\}})$. $\Box$

\subsection{Proof of Lemma \ref{lem:3}}
\emph{Proof:} Consider any set $A$  such that $A \in \mathcal{I}^{R}_S$. Because  $A \in \mathcal{I}^{R}_S $, we have $A\cup S \in \mathcal{I}$ and  $A\subseteq R$. Since  $S'\subseteq  S$ and $\mathcal{I}$ is downward-closed, we have $A \cup S' \in \mathcal{I}$.   Hence, $A \cup S' \in \mathcal{I}^{R}_{S'}$. Therefore,  $\mathcal{I}^{R}_S \subseteq \mathcal{I}^{R}_{S'}$. $\Box$

\subsection{Proof of Lemma \ref{lem:5}}
\emph{Proof:} Consider an arbitrary partial realization $\psi$ such that $\mathrm{dom}(\psi) \in \mathcal{I}$. Define $g(\cdot) = f( \cdot \mid \psi)$. For any partial realization $\psi'$ such that $\psi \subseteq \psi'$ and $\mathrm{dom}(\psi') \in \mathcal{I}$, define $g_{avg}(\pi| \psi') = \mathbb{E}_{\Phi, \Pi}[g(\mathrm{dom}(\psi') \cup V(\pi, \Phi), \Phi)-g(\mathrm{dom}(\psi'), \Phi)\mid \Phi\sim \psi']$. Given any $R\subseteq V$ such that $\mathrm{dom}(\psi') \cap R = \emptyset$, we next show that $g_{avg}(\pi| \psi')=f_{avg}(\pi| \psi')$ for any $\mathcal{I}^{R}_{\mathrm{dom}(\psi')}$-restricted policy $\pi\in \Omega(\mathcal{I}^{R}_{\mathrm{dom}(\psi')})$.
\begin{eqnarray}
&& g_{avg}(\pi| \psi') = \\
&&\mathbb{E}_{\Phi, \Pi}[g(\mathrm{dom}(\psi') \cup V(\pi, \Phi), \Phi)-g(\mathrm{dom}(\psi'), \Phi)\mid \Phi\sim \psi']\nonumber \\
&&= \mathbb{E}_{\Phi, \Pi}[(f(\mathrm{dom}(\psi') \cup V(\pi, \Phi), \Phi)-f(\mathrm{dom}(\psi), \Phi))-(f(\mathrm{dom}(\psi'), \Phi)-f(\mathrm{dom}(\psi), \Phi))\mid \Phi\sim \psi'] \nonumber\\
&&= \mathbb{E}_{\Phi, \Pi}[f(\mathrm{dom}(\psi') \cup V(\pi, \Phi), \Phi)-f(\mathrm{dom}(\psi'), \Phi)\mid \Phi\sim \psi'] \nonumber\\
&&= f_{avg}(\pi| \psi') \label{eq:1}
\end{eqnarray}
The second equality is due to $g(\cdot) = f( \cdot \mid \psi)$. Consider any two partial realizations $\psi^a$ and $\psi^b$ such that $\psi\subseteq \psi^a\subseteq \psi^b$ and $\mathrm{dom}(\psi^b)\in \mathcal{I}$. Recall that $\pi^*(\mathcal{I}^{R}_{\mathrm{dom}(\psi^b)}, \psi^a)$ denotes the optimal  $\mathcal{I}^{R}_{\mathrm{dom}(\psi^b)}$-restricted policy on top of $\psi^a$.  Plugging $\pi^*(\mathcal{I}^{R}_{\mathrm{dom}(\psi^b)}, \psi^a)$ and the partial realization $\psi^a$ into  (\ref{eq:1}), we have  
\begin{eqnarray}
g_{avg}(\pi^*(\mathcal{I}^{R}_{\mathrm{dom}(\psi^b)}, \psi^a)| \psi^a) &=& f_{avg}(\pi^*(\mathcal{I}^{R}_{\mathrm{dom}(\psi^b)}, \psi^a)| \psi^a)\nonumber \\
&\geq& f_{avg}(\pi^*(\mathcal{I}^{R}_{\mathrm{dom}(\psi^b)}, \psi^b)| \psi^b) \nonumber\\
&=& g_{avg}(\pi^*(\mathcal{I}^{R}_{\mathrm{dom}(\psi^b)}, \psi^b)| \psi^b)\nonumber
\end{eqnarray}
Both equalities are due to (\ref{eq:1}), and the inequality is due to $f: 2^{V}\times O^V\rightarrow \mathbb{R}$ is policywise submodular with respect to a prior $p(\phi)$ and an independence system  $(V, \mathcal{I})$. $\Box$

\subsection{Proof of Lemma \ref{lem:1}}
\emph{Proof:} We prove this lemma in two cases.
\begin{enumerate}
\item When $s\in T$, $\pi_T$ first selects $s$, then adopts the optimal $\mathcal{I}^{T\setminus\{s\}}_{\{s\}}$-restricted policy $\pi^*(\mathcal{I}^{T\setminus\{s\}}_{\{s\}},\Phi(s))$ on top of $\Phi(s)$. Let $A$ denote the random set selected by $\pi_T$ after $s$ is being selected. Thus, $A$ is a $\mathcal{I}^{T\setminus\{s\}}_{\{s\}}$-restricted policy. It follows that $A \subseteq T\setminus\{s\}$  and $A\cup \{s\} \in \mathcal{I}$. Hence, $A \cup\{s\} \subseteq T$  and $A\cup \{s\} \in \mathcal{I}$. Thus, $A\cup\{s\}\in \mathcal{I}^{T}_\emptyset$.
\item When $s\notin T$, $\pi_T$ adopts the optimal $\mathcal{I}^{T}_{\{s\}}$-restricted policy $\pi^*(\mathcal{I}^{T}_{\{s\}},\emptyset)$  on top of $\emptyset$. Let $B$ denote the set selected by $\pi_T$. We have $B \in \mathcal{I}^{T}_{\{s\}}$. Because $\mathcal{I}^{T}_{\{s\}} \subseteq \mathcal{I}^{T}_\emptyset$ (Lemma \ref{lem:3}), we have $B \in \mathcal{I}^{T}_\emptyset$. \end{enumerate}$\Box$
    \begin{figure*}[hptb]
\includegraphics[scale=0.9]{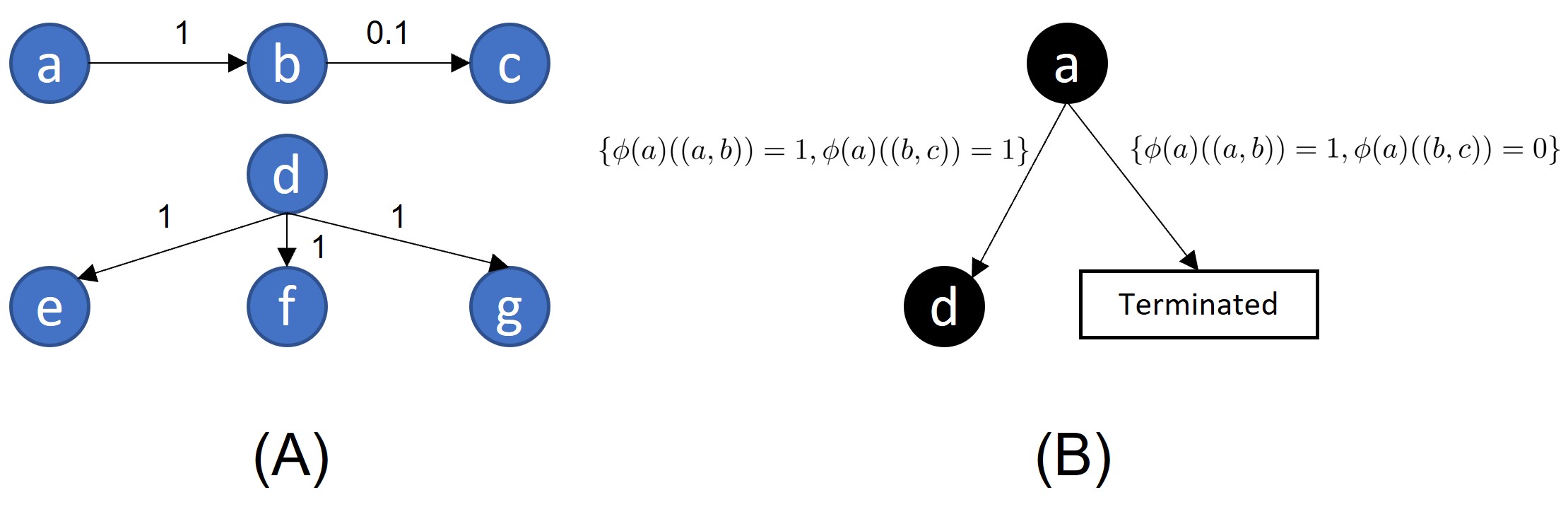}
\caption{(A) A social network with propagation probability on each edge. (B) The decision tree of $\pi$.}
\label{fig:puri_reduction}
\end{figure*}

\subsection{Proof of Lemma \ref{lem:2}}
\emph{Proof:} Because selecting a single item is also a policy, policy-adaptive submodularity
implies adaptive submodularity. We next prove that policy-adaptive submodularity implies policywise submodularity. Consider any two partial realizations $\psi^a$ and $\psi^b$ such that $\psi^a\subseteq \psi^b$ and $\mathrm{dom}(\psi^b)\in \mathcal{I}$, and any $R\subseteq V$ such that $R \cap \mathrm{dom}(\psi^b)=\emptyset$. Because $f: 2^{V}\times O^V\rightarrow \mathbb{R}$  is policy-adaptive submodular with respect to $p(\phi)$, we have
\begin{eqnarray}
f_{avg}(\pi^*(\mathcal{I}^{R}_{\mathrm{dom}(\psi^b)}, \psi^b)| \psi^a) \geq f_{avg}(\pi^*(\mathcal{I}^{R}_{\mathrm{dom}(\psi^b)}, \psi^b)| \psi^b) \label{eq:5}
\end{eqnarray}
Based on the definition of $\pi^*(\mathcal{I}^{R}_{\mathrm{dom}(\psi^a)}, \psi^a)$, we have $f_{avg}(\pi^*(\mathcal{I}^{R}_{\mathrm{dom}(\psi^b)}, \psi^a)| \psi^a) \geq f_{avg}(\pi^*(\mathcal{I}^{R}_{\mathrm{dom}(\psi^b)}, \psi^b)| \psi^a)$.  Together with (\ref{eq:5}), we have $f_{avg}(\pi^*(\mathcal{I}^{R}_{\mathrm{dom}(\psi^b)}, \psi^a)| \psi^a) \geq f_{avg}(\pi^*(\mathcal{I}^{R}_{\mathrm{dom}(\psi^b)}, \psi^b)| \psi^b)$. Hence, $f: 2^{V}\times O^V\rightarrow \mathbb{R}$  is policywise submodular with respect to $p(\phi)$ and any independence system $(V, \mathcal{I})$. $\Box$

\subsection{Proof of Proposition \ref{pro:3}}
 \emph{Proof:} For any partial realization $\psi$, let $\mathrm{dom}(\psi)=\{u|\exists u, v, w, (\phi(u)((v,w))=1) \in \psi \mbox{ or }(\phi(u)((v,w))=0)\in \psi\}$ denote the set of seeds that are selected under $\psi$, $V(\psi)=\{w| w\in \mathrm{dom}(\psi) \mbox{ or }\exists u, v, w, (\phi(u)((v,w))=1)\in \psi \}$ denote the set of all individuals that are activated under $\psi$, and $E(\psi)=\{(v,w)| \exists u, v, w, (\phi(u)((v,w))=1)\in \psi\mbox{ or }(\phi(u)((v,w))=0)\in \psi\}$ denote the set of observed edges under $\psi$. Consider any two partial realizations $\psi^a$ and $\psi^b$ such that $\psi^a\subseteq \psi^b$ and $\mathrm{dom}(\psi^b)\in \mathcal{I}$, and any $R\subseteq V$ such that $R\cap \mathrm{dom}(\psi^b)=\emptyset$. Given an optimal $\mathcal{I}^{R}_{\mathrm{dom}(\psi^b)}$-restricted policy $\pi^*(\mathcal{I}^{R}_{\mathrm{dom}(\psi^b)}, \psi^b)$ and its decision tree \emph{conditioned on $\Phi\sim \psi^b$}, we next build a  $\mathcal{I}^{R}_{\mathrm{dom}(\psi^b)}$-restricted policy $\pi$ such that $f_{avg}(\pi| \psi^a) \geq f_{avg}(\pi^*(\mathcal{I}^{R}_{\mathrm{dom}(\psi^b)}, \psi^b)| \psi^b)$.

\textbf{ Construction of $\pi$.} We first introduce some notations. Let $V'=V\setminus V(\psi^b)$ and $E'=E \setminus E(\psi^b)$. For each $u\in V'$ and a realization $\phi$, we define the $V'$-restricted realization $\theta(\phi)(u)$ of $u$ as a function $\theta(\phi)(u): E' \rightarrow \{1, 0, ?\}$ such that for each edge $(v, w)\in E'$, if $\phi(u)((v,w))=1$ and  $v$  can be reached by $u$ through some path composed of live edges from $E'$, then set $\theta(\phi)(u)((v,w))=1$; if $\phi(u)((v,w))=0$ and $v$ can be reached by $u$ through some path composed of live edges from $E'$, then set $\theta(\phi)(u)((v,w))=0$; otherwise, set $\theta(\phi)(u)((v,w))=?$. Intuitively, given a realization $\phi$, $\theta(\phi)(u)$ contains the statuses of those edges whose statuses can be observed after $u$ is being selected even if we remove all edges in $E(\psi^a)$  from $G$. Let $\theta(\Phi)(u)$ denote a random realization of the $V'$-restricted realization of $u$ and define $\theta(\Phi)=\{\theta(\Phi)(u) \mid u\in V'\}$. Consider a fixed $V'$-restricted realization  $\theta(\phi)$, a full realization $\phi'$, and any individual $u\in V'$, we say $\theta(\phi)(u)$ is consistent with $\phi'(u)$, denoted $\theta(\phi)(u)\sim \phi'(u)$, if they are equal everywhere in the domain of edges whose statuses can be observed under both $\theta(\phi)(u)$ and $\phi'(u)$.

Now we are ready to build $\pi$ based on $\pi^*(\mathcal{I}^{R}_{\mathrm{dom}(\psi^b)}, \psi^b)$. For each pair of a partial realization $\psi$  conditioned on $\Phi\sim \psi^b$ and an individual $e$ such that $\pi^*(\mathcal{I}^{R}_{\mathrm{dom}(\psi^b)}, \psi^b)(\psi) = e$, i.e,  $\pi^*(\mathcal{I}^{R}_{\mathrm{dom}(\psi^b)}, \psi^b)$ selects $e$ after observing $\psi$, we define $\pi(\psi')=e$ for all $\psi'$ such that  $\mathrm{dom}(\psi)=\mathrm{dom}(\psi')$ and for all $(u, \phi(u)) \in \psi$ and $(u, \phi'(u)) \in \psi'$,  $\theta(\phi')(u)\sim \phi(u)$. The intuition behind $\pi$ is to mimic the execution of  $\pi^*(\mathcal{I}^{R}_{\mathrm{dom}(\psi^b)}, \psi^b)$ conditioned on $\Phi\sim \psi^b$ by ignoring the feedback from those individuals which can not be activated through a path composed of live edges in $E'$.

 We first show that $\pi$ is a  $\mathcal{I}^{R}_{\mathrm{dom}(\psi^b)}$-restricted policy.  Recall that for any partial realization $\psi'$ and individual $e$ such that $\pi(\psi') = e$, there must exist some $\psi$ such that    $\pi^*(\mathcal{I}^{R}_{\mathrm{dom}(\psi^b)}, \psi^b)(\psi) = e$ and $\mathrm{dom}(\psi)=\mathrm{dom}(\psi')$. Because $\pi^*(\mathcal{I}^{R}_{\mathrm{dom}(\psi^b)}, \psi^b)$ is a $\mathcal{I}^{R}_{\mathrm{dom}(\psi^b)}$-restricted policy, we have $\{e\}\cup \mathrm{dom}(\psi) \in  \mathcal{I}^{R}_{\mathrm{dom}(\psi^b)}$ and $e\in R$. This indicates that $\{e\}\cup \mathrm{dom}(\psi') \in  \mathcal{I}^{R}_{\mathrm{dom}(\psi^b)}$ and $e\in R$ due to $\mathrm{dom}(\psi)=\mathrm{dom}(\psi')$. Thus, $\pi$ is a  $\mathcal{I}^{R}_{\mathrm{dom}(\psi^b)}$-restricted policy.

%

 We next prove that $f_{avg}(\pi| \psi^a) \geq f_{avg}(\pi^*(\mathcal{I}^{R}_{\mathrm{dom}(\psi^b)}, \psi^b)| \psi^b)$.
Following the design of $\pi$, $\pi$ and $\pi^*(\mathcal{I}^{R}_{\mathrm{dom}(\psi^b)}, \psi^b)$ select the same group of individuals as seeds conditioned on any $V'$-restricted realization $\theta(\phi)$. Thus, for any node $u\in V'$, if $u$ is activated by $\pi^*(\mathcal{I}^{R}_{\mathrm{dom}(\psi^b)}, \psi^b)$ conditioned on $\psi^b$ and   any $V'$-restricted realization $\theta(\phi)$, which implies that there exists a live path in $E'$ such that it connects $u$ with some seed selected by $\pi^*(\mathcal{I}^{R}_{\mathrm{dom}(\psi^b)}, \psi^b)$, then $u$ must be activated by $\pi$ conditioned on $\psi^a$ and $\theta(\phi)$. Hence, the marginal utility of $\pi$ with respect to $\psi^a$ is no less than the marginal utility of $\pi^*(\mathcal{I}^{R}_{\mathrm{dom}(\psi^b)}, \psi^b)$ with respect to $\psi^b$, i.e., \begin{eqnarray}&&\mathbb{E}[f_{avg}(\pi| \psi^a)| \Phi\sim \theta(\phi) \cup \psi^a] \nonumber\\
&&\geq \mathbb{E}[f_{avg}(\pi^*(\mathcal{I}^{R}_{\mathrm{dom}(\psi^b)}, \psi^b)| \psi^b)|  \Phi\sim  \theta(\phi) \cup \psi^b] \label{eq:disapoint}
 \end{eqnarray}
  Thus, \begin{eqnarray}
&&f_{avg}(\pi| \psi^a) =  \sum_{\theta(\phi)} \Pr[\theta(\Phi)=\theta(\phi)|\psi^a]\mathbb{E}[ f_{avg}(\pi|  \Phi\sim \theta(\phi)\cup \psi^a)] \nonumber\\
 &&\geq\sum_{\theta(\phi)} \Pr[\theta(\Phi)=\theta(\phi) |\psi^b] \mathbb{E}[f_{avg}(\pi^*(\mathcal{I}^{R}_{\mathrm{dom}(\psi^b)}, \psi^b)| \psi^b)| \Phi\sim \theta(\phi)\cup \psi^b]\nonumber\\
 &&=  f_{avg}(\pi^*(\mathcal{I}^{R}_{\mathrm{dom}(\psi^b)}, \psi^b)| \psi^b) \label{eq:9}
 \end{eqnarray}
The inequality is due to   (\ref{eq:disapoint}) and the fact that $\theta(\Phi)$ has the same distribution conditioned on both $\psi^a$ and $\psi^b$ since $\psi^a\subseteq \psi^b$.

 Because $\pi^*(\mathcal{I}^{R}_{\mathrm{dom}(\psi^b)}, \psi^a)$ is the optimal $\mathcal{I}^{R}_{\mathrm{dom}(\psi^b)}$-restricted policy on top of $\psi^a$, we have $f_{avg}(\pi^*(\mathcal{I}^{R}_{\mathrm{dom}(\psi^b)}, \psi^a)| \psi^a)\geq f_{avg}(\pi| \psi^a)$. Together with (\ref{eq:9}), we have
 \begin{eqnarray}
f_{avg}(\pi^*(\mathcal{I}^{R}_{\mathrm{dom}(\psi^b)}, \psi^a)| \psi^a) \geq f_{avg}(\pi^*(\mathcal{I}^{R}_{\mathrm{dom}(\psi^b)}, \psi^b)| \psi^b) \label{eq:8}
\end{eqnarray} $\Box$

\subsection{Proof of Lemma \ref{lem:7}}
  \emph{Proof:}  We construct an example to show that  (\ref{eq:11}) is not policy-adaptive submodular. Consider a social network in Figure \ref{fig:puri_reduction}(A) and a policy $\pi$ listed in Figure \ref{fig:puri_reduction}(B). We first analyze the expected marginal utility of $\pi$ on top of a partial realization $\psi^b=\{\phi(b)((b,c))=1\}$. Following the policy $\pi$, we select individuals $a$ and $d$ given  $\psi^b=\{\phi(b)((b,c))=1\}$. Hence,
\begin{eqnarray}\label{eq:6}
f_{avg}(\pi | \psi^b) = 7-2 = 5
\end{eqnarray}

We next analyze the expected marginal utility of $\pi$ on top of another partial realization $\psi^a=\emptyset$.  In the case of $\phi(a)((b,c))=0$, $\pi$ selects $a$ only. In the case of $\phi(a)((b,c))=1$, $\pi$ selects $a$ and $d$. Hence, $f_{avg}(\pi | \psi^a)$ is
\begin{eqnarray}\label{eq:7}
(\Pr[\phi(a)((b,c))=0]\times 2 + \Pr[\phi(a)((b,c))=1]\times 7) - 0 = 2.5
\end{eqnarray}
(\ref{eq:6}) and (\ref{eq:7}) imply that $f_{avg}(\pi | \psi^b) > f_{avg}(\pi | \psi^a)$.

The above example, together with Lemma \ref{lem:2} and  Proposition \ref{pro:3}, implies this lemma. $\Box$




\end{document}